\documentclass[10pt,twocolumn,letterpaper]{article}

\usepackage{iccv}
\usepackage{times}
\usepackage{epsfig}
\usepackage{graphicx}
\usepackage{amsmath}
\usepackage{amssymb}
\usepackage{subfigure}
\usepackage{multirow}
\usepackage{algorithm}
\usepackage{algpseudocode}
\usepackage{pdfpages}

\usepackage[breaklinks=true,bookmarks=false]{hyperref}
\usepackage{cleveref}

\iccvfinalcopy 

\def\iccvPaperID{****} 

\ificcvfinal\fi

\begin{document}

\title{Fine-Grained is Too Coarse: \\A Novel Data-Centric Approach for Efficient Scene Graph Generation}


\author{Maëlic Neau$^{1,2}$ \hspace{0.3in} Paulo E. Santos$^1$ \hspace{0.3in} Anne-Gwenn Bosser$^2$ \hspace{0.3in} Cédric Buche$^{2}$\\
$^1$College of Science and Engineering, Flinders University, Australia\\
$^2$Ecole Nationale d'Ingénieurs de Brest, France
\\{\tt\small \{neau, buche, bosser\}@enib.fr,
paulo.santos@flinders.edu.au }
}

\maketitle
\ificcvfinal\thispagestyle{empty}\fi

\begin{abstract}

Learning to compose visual relationships from raw images in the form of scene graphs is a highly challenging task due to contextual dependencies, but it is essential in computer vision applications that depend on scene understanding. 
However, no current approaches in Scene Graph Generation (SGG) aim at providing useful graphs for downstream tasks. Instead, the main focus has primarily been on the task of unbiasing the data distribution for predicting more fine-grained relations. That being said, all fine-grained relations are not equally relevant and at least a part of them are of no use for real-world applications. In this work, we introduce the task of Efficient SGG that prioritizes the generation of relevant relations, facilitating the use of Scene Graphs in downstream tasks such as Image Generation. To support further approaches, we present a new dataset, VG150-curated, based on the annotations of the popular Visual Genome dataset. We show through a set of experiments that this dataset contains more high-quality and diverse annotations than the one usually use in SGG. Finally, we show the efficiency of this dataset in the task of Image Generation from Scene Graphs\footnote{Source code: \url{https://github.com/Maelic/VG_curated}}. 

\end{abstract}

\section{Introduction}

The task of Scene Graph Generation (SGG) aims at creating a symbolic representation of a scene by inferring relations between entities as a graph structure. Typically, approaches in SGG rely on detecting object features from an image and then inferring relation predicates between object pairs as \(<subject, predicate, object>\) triplets. Connections between pairs of triplets form a directed acyclic graph in which each vertex refers to an object and its associated image region. Due to its efficient representation capacity, this task holds strong promises for other downstream tasks such as Image Captioning \cite{xu2019scene, wang2019role, yang2022reformer} or Visual Question Answering \cite{damodaran2021understanding, lee2019visual}. Recent contributions to the field highlight an opportunity for SGG to support the reasoning of an embodied agent by leveraging both the spatial and semantic latent context of a scene in a single representation \cite{amodeo2022og, li2022embodied, amiri2022reasoning}. However, despite a vast amount of work in the last few years, the performance of the best approaches is far from optimal, and the usage in downstream tasks is limited \cite{zhu_scene_2022}.
A set of problems have been raised by the community to explain this slow pace, the main one is the long-tail distribution of predicates \cite{tang_unbiased_2020,yu_cogtree_2021,li_bipartite_2021}. In fact, due to annotation biases, datasets used in SGG tend to have more annotated samples with vague predicates (e.g. \texttt{on}, \texttt{has} or \texttt{near}) rather than with fine-grained ones (e.g. \texttt{riding}, \texttt{under} or \texttt{eating}). 
\begin{figure}[]
    \centering
        \subfigure[]{
        \centering
            \includegraphics[width=0.35\columnwidth]{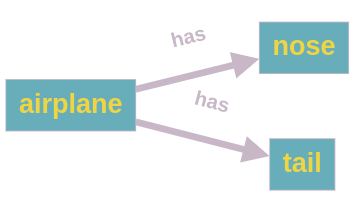}\label{fig:plane_graph1}
        } 
        \subfigure[]{
        \centering
            \includegraphics[width=0.35\columnwidth]{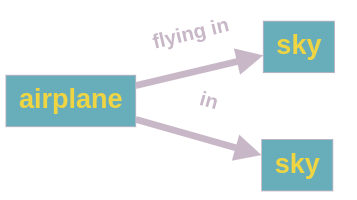}\label{fig:plane_graph2}
        } 
        \subfigure[]{
        \centering
            \includegraphics[width=0.35\columnwidth]{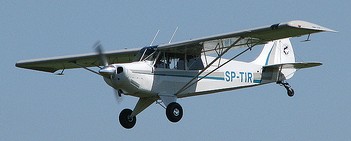}\label{fig:plane}
        }
    \caption{Top-2 relations predictions for \Cref{fig:plane}: \ref{fig:plane_graph1} is using original dataset  and \ref{fig:plane_graph2} is using our curated dataset. Note that more relevant relations are obtained in the latter.}
    \label{fig:plane_predictions}
\end{figure}
While this issue has been largely investigated under the name of \textbf{Unbiased SGG} \cite{ye_linguistic_2021, yu2017visual, dong_stacked_2022, zhong_learning_2021,yoon_unbiased_2022, gu2019scene, suhail_energy-based_2021, li_bipartite_2021}, other aspects of the task have been left aside, such as the amount of actual useful information conveyed by a scene graph structure. Inspired by recent approaches in this direction \cite{wang2020sketching}, we introduce the task of \textbf{Efficient SGG} that aims at extracting the maximum quantity of \textit{valuable information} from an input scene, in contrast to current approaches that focus on extracting \textit{fine-grained information} first. This new approach is highly beneficial to downstream tasks where predicting major events from the scene is more important than predicting detailed but minor ones (see a comparison in \Cref{fig:plane_predictions}).

To support efficient learning for this task, we focus on providing a novel high-quality dataset by leveraging the existing but noisy annotations from the largest and main used dataset in SGG, Visual Genome (VG) \cite{krishna2017visual}. In contrast to other curation approaches of VG \cite{liang2019vrr, wang2020sketching, plesse2020focusing}, we focus on preserving the semantics conveyed by Scene Graph structures during pruning while annotations that are irrelevant for downstream tasks are removed, thus creating an optimised dataset for the task of Efficient SGG. To demonstrate the necessity of our approach, we show that SGG models trained on the current version of Visual Genome are inefficient in downstream tasks: first, they are biased toward predicting irrelevant relations with overconfidence. Second, the poor connectivity in annotated samples penalises the learning process resulting in low-quality graphs. Our approach tackles these two problems, resulting in a new high-quality dataset for the task that improves the performance of baseline models by a strong margin. We further evaluate the use of this new dataset in the task of Image Generation. 

The main contributions of this paper are threefold: \textbf{(i)} a study on the impact of sample connectivity and irrelevant relations in the learning process of baseline models in SGG ;
\textbf{(ii)} a new curation process of the Visual Genome dataset that results in VG150-curated, a new high-quality dataset for SGG ;
\textbf{(iii)} a study on the benefits of VG150-curated in SGG and downstream tasks.

\section{Related Work}

Since the first description of the task \cite{xu2017scene}, SGG has drawn widespread attention in the computer vision and natural language processing communities. Popular approaches combine object detection backbones such as the popular Faster-RCNN \cite{ren2015faster} with a graph generation model in a two-stage pipeline \cite{xu2017scene, zhang2017visual, tang2019learning, li_bipartite_2021, lin_gps-net_2020, gu2019scene, wang_exploring_2019}.  However, concerns about biases in large-scale datasets such as Visual Genome \cite{krishna2017visual} have been rapidly raised, resulting in the task of Unbiased SGG \cite{wen_unbiased_2020, tang_unbiased_2020, yoon_unbiased_2022, yan2020pcpl, yu_cogtree_2021, dong_stacked_2022, desai2021learning}. The idea is to improve predicate prediction using different model-agnostic techniques and training strategies such as the Total Direct Effect \cite{tang_unbiased_2020} or probability distribution \cite{li2022ppdl} and evaluate them on a set of baseline models \cite{zellers2018neural, tang2019learning, xu2017scene}. On the other hand, new approaches \cite{cong_reltr_2022, li2022sgtr, lu2021context, liu2021fully} are considering the task in a one-stage fashion, learning relationships from image features directly. Still, these approaches are assuming that all relations are equal and share the same amount of information on the scene in the learning process. This creates models that extensively predict meaningless relations with high confidence, hindering the performance of downstream tasks that depend on relevant predictions. A first step towards solving this issue is to enforce \textit{relevant} annotations in data-centric approaches \cite{li2022devil, zhang2022fine}. 

\subsection{Data-Centric Approaches in SGG}

To the best of our knowledge, only a few current approaches are considering the Visual Genome dataset biases from a data-centric perspective. In VrR-VG \cite{liang2019vrr}, the authors based their assumption on the fact that relations that can be easily inferred with only spatial information from an object pair (i.e. bounding box coordinates) are not visually relevant. This results in the removal of common relations and leads to sparse annotations where only rare and very specific relations are annotated for which the use in downstream tasks is very limited. 
Other approaches are focusing on balancing the predicate distribution \cite{plesse2020focusing} or filtering similar or vague predicates \cite{abou2022topology} to improve the relevance of the annotations. However, these methods assume a consistent use of the same predicate across the annotations which is not true due to the inherent \textit{semantic ambiguity} of natural language \cite{zhang2022fine}.
Thus, we believe that curating the annotations based on the predicate distribution alone is not a viable strategy. Intuitively, taking into account the triplets' distribution seems to alleviate this semantic ambiguity and could be a more beneficial strategy.

Recent work has focused on re-sampling and de-noising the Visual Genome dataset. 
Different techniques were used to improve the quality of annotations such as internal and external transfer \cite{zhang2022fine} or clustering strategies \cite{li2022devil}. The reported results showed a non-negligible impact on the training of baseline models for SGG (by up to 25.2\% in certain cases \cite{li2022devil}). In fact, looking at those results \cite{zhang2022fine}, it is possible to conclude that at this stage cleaning the dataset is more beneficial for the task than implementing new models.
Finally, research reported in \cite{zhang2017visual}, \cite{dai_detecting_2017}, \cite{abdelkarim2021exploring}, and \cite{xu2017scene} are splitting the original annotations based on different object and predicate classes frequency. These approaches were not filtering relationships to explicitly address inherent biases, here the data split addressed only minor annotation issues such as object class de-noising \cite{zhang2017visual}. A comparison of the different splits of Visual Genome reported in SGG papers is available in \Cref{tab:datasets}.  
\begin{table}[]
    \centering
    \begin{tabular}{c|c|c|c}
        \hline
         \bf VG Split &  \bf \#Images &  \bf \#Objects &  \bf \#Predicates \\ 
         \hline 
         VG80K \cite{zhang2019large} & 104,832 & 53,304 & 29,086 \\
         \hline
         VG150 \cite{xu2017scene} & 105,414 & 150 & 50 \\
         \hline
         VrR-VG \cite{liang2019vrr} & 58,983 & 1,321 & 117 \\
         \hline
    \end{tabular}
    \caption{Comparison of the main splits of Visual Genome used in SGG. \#Images is the total number of annotated images, \#Objects and \#Predicates count the number of classes.}
    \label{tab:datasets}
\end{table}
In contrast to prior work, the present paper focuses on building a data split that encompasses only visually-relevant annotations to support usage in downstream tasks. 
To do so, we introduce a new definition of visually-relevant relation based on the following assumption: {\em a relation is not relevant if it describes a composition between parts of an entity that is true in a general sense and that could be inferred using external knowledge (e.g. $<man, has, arm>$)}.

This definition of relevant relations is not related to the nature of the dataset but models instead the semantic information conveyed by a scene graph structure.
Moreover, we also consider the issue of connectivity of the annotated samples which has never been addressed before. The next section details these issues, while also arguing why their solution causes a positive impact on the downstream tasks.

\section{Problem Definition}

The Visual Genome dataset \cite{krishna2017visual} is the largest and most widely adopted dataset for SGG. Its annotations have been collected by annotators in the form of region captions. Then, different parsing techniques have been applied to retrieve $<subject, predicate, object>$ triplets for each region. Because annotators were not constrained to use any particular vocabulary, this process resulted in more than 53K object classes and 27K predicate classes, where more than 50\% of them only have one sample. This split of the data is usually referred to as VG80K \cite{zhang2019large}. To support efficient learning for SGG, the common practice is to prune annotations from VG80K, keeping only a selection of the top-$k$ predicate and object classes. However, when doing so, no current approaches are aiming at preserving the graph structure as well as keeping the relevant information about the scene. In this work, we present an approach that is able to extract annotations for $k$ object and predicate classes while ensuring to preserve most of the information from the original samples. We believe that this could be achieved by \textbf{(1)} preserving the connectivity of the original graph and \textbf{(2)} extensively pruning irrelevant annotations.

\subsection{Connectivity}

This work uses the following notation: a graph \(G=(V,E)\) represents all relations in a given image with a set of edges \(E\) and a set of vertices \(V\). It is important to notice here that \(G\) is not necessarily fully connected, as some vertices or edges could have been removed from the original annotations. 
We denote the average graph size on a set of $n$ graphs as: \(\bar{s} = \frac{1}{n} \sum_{i=1}^{n} |E(G_i)|.\)

\begin{figure}[]
  \centering
  \includegraphics[width=\linewidth, height=12em]{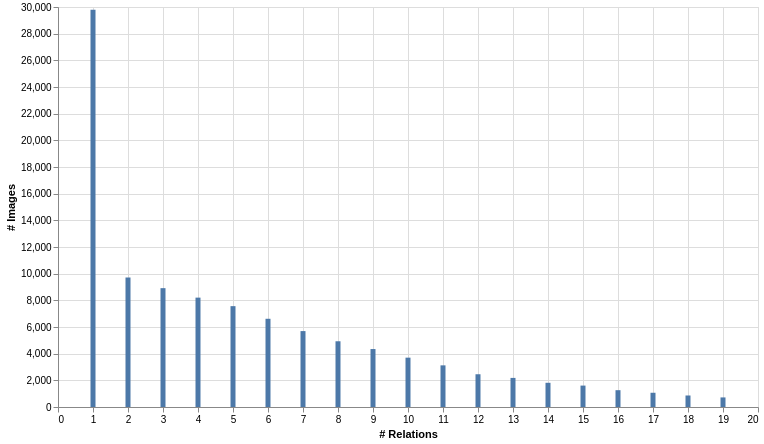}
  \caption{Graph size $\bar{s}$ per image in VG150.}
  \label{fig:vg150_image_distr}
\end{figure}

The average graph size in the original annotations of Visual Genome is high, with $\bar{s}=19.02$. However, when pruning the dataset to keep only the top-$k$ object and predicate classes, a large number of annotations were removed leading to $\bar{s}=6.98$ in the VG150 split \cite{xu2017scene}. \Cref{fig:vg150_image_distr} shows the number of relations with respect to the number of images in VG150, where we can see that the distribution of graph size is long-tailed with more than 28\% of samples that only contain one relation. The average vertex degree $d(v)$ is also low, with an average of 2.02 against 2.34 for VG80K.
These figures can easily be explained by the applied pruning strategy which selected object and predicate classes based only on their overall frequency over the dataset. We believe that this negatively affects the performance of SGG approaches, especially methods that explicitly model the context of every relation using, for instance, Iterative Message Passing \cite{xu2017scene} or bipartite matching \cite{li_bipartite_2021}. This is solved in this work by selecting annotated samples based on their inter-connectivity rather than overall frequency.

\subsection{Irrelevant Relationships}

Besides a connectivity issue, the annotations of Visual Genome are also biased with the over-representation of certain triplets. In fact, we observed that some invariant relations (such as $<man, has, head>$) are over-represented in the dataset, creating a bias for models that will always select those relations with over-confidence compared to others. Because the current evaluation metrics Recall@k and meanRecall@k \cite{tang2019learning} are ranking metrics, this leads to poor performance of the task. Previous work \cite{zellers2018neural} enumerated 3 different relation categories in Visual Genome, as follows: geometric, possessive, and semantic (i.e. actions or activities). The geometric category represents spatial relations such as $<cup, on, table>$; possessive relations are composed of an entity and an artifact such as $<car, has, wheel>$; finally, semantic relations represent activities or actions such as $<man, riding, bike>$. In this context, we categorised the top 50 triplets to see if we can experience a certain pattern in over-represented relations.
\begin{figure}
    \centering
    \includegraphics[width=\linewidth]{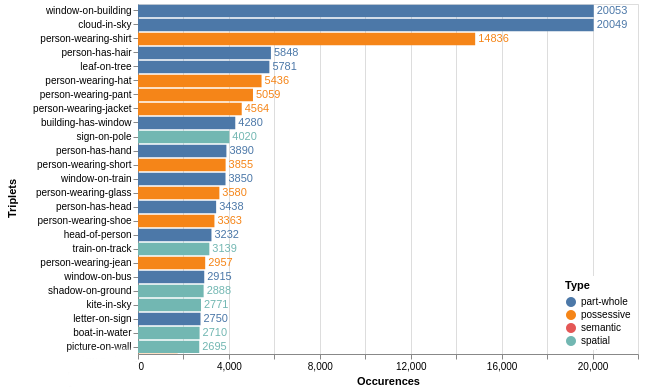}
    \caption{Distribution of the top 25 relations in VG150.}
    \label{fig:vg80k_distri}
\end{figure}
We followed the same classes as in \cite{zellers2018neural} except that we made a distinction between invariant possessive relation (i.e. \textit{part-whole}) and possessive attributes such as clothing (denoted as the \textit{possessive} category here). \Cref{fig:vg80k_distri} shows the number of relations with respect to the number of images in VG150, where we can see that part-whole relations were prevalent with 55.1\% of the total number of occurrences for the top 25 more represented relations. 
As explained in \cite{liang2019vrr, zhu_scene_2022, wang2020sketching}, these relations may be biasing the learning process because they are true in the general sense and do not depend on visual features of the scene. We call this \textit{invariant relationship bias}. In order to verify this assumption, we conducted an experiment on predictions obtained by the Motifs-TDE model \cite{tang_unbiased_2020} on the test set of the VG150 dataset. Results are shown in \Cref{fig:ratio} where we can see that part-whole relations are overly predicted in comparison to the ground truth annotations.
This issue matters in regard to the importance of each relation in the global context of the scene. For instance, it is questionable how possessive relations such as \(<man, has, head>\) are actually relevant to describe the scene, especially for usage in downstream tasks such as Visual-Question Answering where the amount of noise in the predicted visual relations is critical. In the next section, we detail our approach to solving these two issues by introducing two novel curation methods.

\section{Data Curation}\label{curation}

We started with the original version of the Visual Genome dataset that is pre-processed to clean the annotations as described in the sequence. For the object regions, we replicated the approach proposed by \cite{xu2017scene} to merge bounding boxes with an Intersection over Union (IoU) greater than or equal to $0.9$. For the textual annotations, we also followed \cite{xu2017scene} to remove stop-words and punctuation using the alias dictionaries provided by the authors of the dataset \footnote{\scriptsize{\url{http://visualgenome.org/api/v0/api_home.html}}}. Finally, we merged synonyms of object classes using WordNet synsets. This process resulted in the \textbf{VG80K} version of the dataset that contains 104,832 images annotated and that is similar to the one introduced in \cite{zhang2019large}. 
In \Cref{connectivity}, we introduce a simple algorithm to improve the number of connected regions. To address the issues of relevance of relations, we focused on categorising and removing irrelevant relations, as detailed in \Cref{filter_part_whole}. 

\begin{figure}
    \centering
    \includegraphics[width=.9\columnwidth, height=11em]{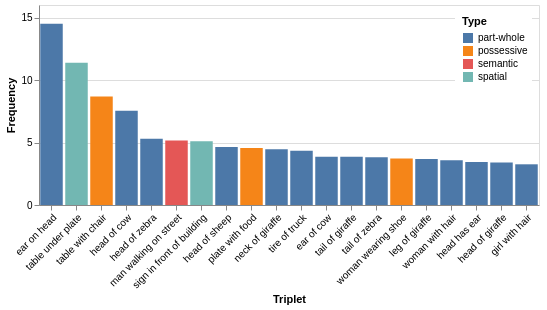}
    \caption{Ratio of predicted triplets over ground truth ones on the test set of VG150. For clarity, we show only the top 20 triplets with more than 20 occurrences. The top-1 triplet is predicted 14,5 more times than the actual ground truth.}
    \label{fig:ratio}
\end{figure}

\subsection{Connectivity}\label{connectivity}

Finding the most connected object ($\textbf{\^o}$) and  predicate ($\textbf{\^p}$) classes for a set of $n$ graphs can be represented as:
\begin{equation}
    \theta(\textbf{\^o}, \textbf{\^p}) = \max_{\textbf{\^o}, \textbf{\^p}} \sum_{k=1}^{n} |G(u,v,w)|, w \in \textbf{\^p} \vee [u, v] \in \textbf{\^o}
\end{equation}

\begin{figure}[]
  \centering
  \includegraphics[width=\linewidth, height=12em]{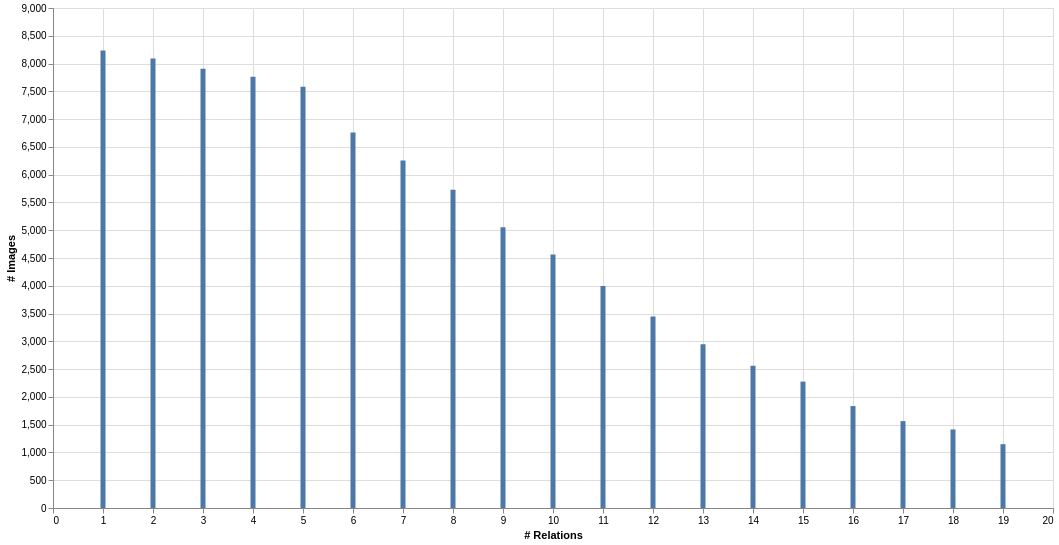}
  \caption{Graph size per number of images in VG150-con.}
\label{fig:vg150_con_distri}
\end{figure}

To  be consistent with VG150, we chose $|\textbf{\^o}|=150$ and $|\textbf{\^p}|=50$. As this is a complex optimisation problem, a satisfying solution can be found by first optimising $\theta(\textbf{\^o})$ and then $\theta(\textbf{\^o}, \textbf{\^p})$ with a fixed set of classes $\textbf{\^o}$. We applied this method to the original data and obtained a new split that we call VG150-con. This split possesses a significantly higher number of relations (22\% more than VG150), with an average graph size $\Bar{s}$ of 8.37 versus 6.98 for VG150, see \Cref{tab:statistics}.  \Cref{fig:vg150_con_distri} displays the top-20 distribution of graph size per image. By comparing this distribution with the original one in \Cref{fig:vg150_image_distr}, we see a clear improvement of the long-tail distribution, proving that our method results in a more connected dataset than the original. More interestingly, we see a net improvement in the average vertex degree (see \Cref{tab:statistics}), moving up from 2.02 to 2.2. This shows that relations are also more interdependent and thus should benefit the context learning of SGG models.  
We further analysed the performance of SGG models on this new split of the data, as presented in \Cref{expe}.

\subsection{Irrelevant Relationships} \label{filter_part_whole}

Building upon our new classification of relevant relations, we employed a new approach to filter the dataset from \textit{part-whole} triplets. To filter categories of relations, previous approaches rely on handcrafted predicate categories \cite{zellers2018neural}.
However, this categorisation only takes into account the semantics of predicates, which suppose that annotations are consistent in the dataset. This assumption is wrong, given the semantic ambiguity introduced in \cite{zhang2022fine}. We give a clear example given the following two triplets from Visual Genome:
\begin{align}
 & man \xrightarrow[]{\text{has}} nose \label{eq:1}\\
&  man \xrightarrow[]{\text{has}} \mathit{surfboard}\label{eq:2}
\end{align}
\noindent When we look at image samples that contain these relations, we see that Formula \ref{eq:1} refers to a \textit{part-whole} relation that would be described as \textit{nose is part of man} while in Formula \ref{eq:2}, the relation is \textit{semantic} and could be described as \textit{man is carrying surfboard}, even though they share the same predicate \textit{has}. 
On the other hand, it has been noticed that there is a strong correspondence between the knowledge embedded in the Visual Genome annotations and in linguistic commonsense knowledge sources such as ConceptNet \cite{gu2019scene,kan2021zero}. Thus, instead of manually labeling every triplet in VG, we chose to compare triplets annotations with a subset of ConceptNet \cite{speer2017conceptnet} that contains only part-whole relations. If a relation has a significant similarity with one from ConceptNet, then we can filter all its occurrences from the original data.
We used the \textit{part-whole} subset of ConceptNet, following the ontology introduced in \cite{ilievski2021dimensions} with the relations 'PartOf', 'HasA', and 'MadeOf'.
Then, we used different approaches to categorize relations between part-whole and non-part-whole from textual annotations only. To evaluate the performance of this filtering, we manually annotated a subset of 1000 random relations from Visual Genome. First, we evaluated the filtering using lexical similarity between \(<subject, object>\) pair in Visual Genome and ConceptNet.
Second, we compared the average of \(<subject, predicate, object>\) Glove embeddings with those from ConceptNet using the cosine similarity. Third, we used different pre-trained Sentence Transformers \cite{reimers-2019-sentence-bert} models to generate sentence similarity embeddings. Finally, we compared those approaches with the predicate-only classification proposed by \cite{zellers2018neural} in which 50 predicate types were classified within semantic, geometric, and possessive classes. 
Results displayed in \Cref{tab:filtering} show that the classification by \cite{zellers2018neural} resulted in the lowest score, this was due to the inconsistency in predicate annotations, as explained before. Sentence-Transformers approaches, as they have been pre-trained on a large corpus of texts, are able to easily generalized and give the best performance. In the choice of embeddings, we prioritized precision over recall as we do not want to discard anything else than \textit{part-whole} relations.
The \textit{all-mpnet-base-v2} \footnote{\scriptsize{\url{https://huggingface.co/sentence-transformers/all-mpnet-base-v2}}} model has shown the best performance in the task, giving satisfactory trade-off between precision and F1 score. This result is consistent with previous work as this model is ranked 5 in the task of Sentence Similarity \footnote{\scriptsize{\url{https://huggingface.co/spaces/mteb/leaderboard} \textit{accessed on the 21/11/2022.}}}.

\begin{table}[]
    \centering
     \begin{tabular}{p{0.5\linewidth} | p{0.1\linewidth}| p{0.15\linewidth}| p{0.065\linewidth}} 
     \hline
     \bf Method &  \bf Recall &  \bf Precision &  \bf F1 \\ 
     \hline
     Predicate only \cite{zellers2018neural} & 0.43 & 0.62 & 0.51 \\ \hline
     Lexical similarity & 0.81 & 0.53 & 0.64\\  \hline
     Glove 6B 300d \textit{(cos=0.7)} & \bf 0.88 & 0.5 & 0.64  \\  \hline
     RoBERTa-large-v1 \textit{(cos=0.7)} & 0.75 & 0.58 & 0.66 \\ \hline
     MiniLM-L6-v2 \textit{(cos=0.7)} & 0.74 & 0.67 & 0.7 \\ \hline
     MpNet-base-v2 \textit{(cos=0.75)} & 0.64 & \bf 0.83 & \bf 0.72 \\ 
     \hline
    \end{tabular}
    \caption{Part-whole relations filtering by comparing with ConceptNet, evaluation on a set of 1000 random samples.}
    \label{tab:filtering}
\end{table}

Using the embeddings produced by \textit{all-mpnet-base-v2}, we were able to extract 36,777 part-whole relations for a total of 416,318 occurrences in VG80K (18\% of the annotations). Before removing those annotations from the original samples, we ensured that no other types of relationships were dependent on them. This step is important because by removing some part-whole relations we could lose some of the semantics of the scene. For instance, the  sub-graph: 
\begin{equation}
 person \xrightarrow[]{\text{has}} hand \xrightarrow[]{\text{holding}} cup
\end{equation}
 describes a semantic relation between the entity \texttt{person} and \texttt{cup}, even if the relation $<person, has, hand>$ is classified as a part-whole relation by our method. In this case, the method proposed in this work can be applied as follows: we added a set of weights $ w:E \rightarrow{\mathbb{R}} $ to the original graph $G=(V,E)$ such as $w=1$ if the edge is a part-whole relation and $0$ otherwise. Given this graph, we performed a pruning strategy that iterates through all edges and removed those that were only dependent on other part-whole relations. This removed from the graph relations deemed as irrelevant to the context of the scene.

Finally, we leverage the strategy employed in \Cref{connectivity} to select the most connected object and predicate classes from this new filtered data. 
This process resulted in a data split with 636,175 filtered relationships, that we call VG150-curated (or VG150-cur). In \Cref{tab:statistics} we see that this split possesses fewer samples than VG150-connected. This is the case because we noticed that in the original dataset, part-whole relations are highly connected with each other, i.e. a relation like $<person, has, head>$ will often be associated with $<head, has, hair>$. While other types of relations are more context-dependent, it is harder to find a set of 150 object and 50 relation classes that are highly connected. However, from \Cref{tab:statistics} we see that VG150-curated still possesses a higher average vertex degree than VG150, proving that our method is efficient to select inter-dependent relations. VG150-curated also possesses a significantly higher number of triplets, showing that the relationships represented are more diverse. 
Without invariant relations, this new split represents a more informative and natural description of scenes. The performance of SGG models with this new split is analysed in the next section.

\begin{table}[]
    \centering
    \resizebox{0.9\columnwidth}{!}{%
    \begin{tabular}{c|cccc}
        \hline
         & \multicolumn{4}{c}{\bf Statistics} \\
         \bf Datasets & \(\bar{d}(v)\) & \(\Bar{s}(G)\) & \#Rels & \#Triplets \\
         \hline
         VG80K & 2.34 & 19.02 & 2,316,063 & 514,526 \\
         VG150 & 2.02 & 6.98 & 622,705 & 35,412 \\
         \hline
         \hline
         VG150-con & 2.20 & 8.38 & 799,412 & 44,851  \\
         VG150-cur & 2.12 & 7.14 & 636,175 & 41,164 \\
         \hline
    \end{tabular}
    }
    \caption{Graph's connectivity and size of the different splits; where $\bar{d}(v)$ represents the average vertex degree; $\Bar{s}(G)$ the average graph size; \#Rels is the total number of relations samples, and \#Triplets is the number of different triplets.}
    \label{tab:statistics}
\end{table}

\begin{table*}[]
    \centering
    \begin{tabular}{c|c| c c c c}
        \hline
         & &  \bf PredCls & \bf SGCls & \bf SGGen & \bf Improv. \\
         \bf Models & \bf Dataset & \bf mR@20/50/100  & \bf mR@20/50/100  & \bf mR@20/50/100 & \bf (avg.) \\ 
         \hline
         \multirow{3}{*}{IMP \cite{xu2017scene}} & VG150 * & 8.8/10.80/11.62 & 4.63/5.82/6.42 & 2.76/4.02/5.0 & \bf - \\
         & VG150-con & 8.8/11.9/13.35 & 5.63/6.76/7.16 & 2.59/4.26/5.61 & \bf $\uparrow$ 10.3\% \\
         & VG150-cur & 9.61/12.61/13.92  & 6.99/8.74/9.44 & 4.09/6.21/7.41 & \bf $\uparrow$ 28\% \\
         \hline
        \multirow{3}{*}{Motifs-TDE \cite{tang_unbiased_2020}} & VG150 & 18.5/25.5/29.1 & 9.8/13.1/14.9 & 5.8/8.2/9.8 & \bf - \\
        & VG150-con & 20.38/28.76/34.06 & 10.3/14.6/17.25 & 8.15/11.53/13.15 & \bf $\uparrow$ 16.1\% \\
         & VG150-cur & 21.38/30.90/36.58 & 13.75/18.55/21.54 & 10.49/14.28/17.10 & \bf $\uparrow$ 37\% \\
         \hline
         \multirow{3}{*}{VCTree-TDE \cite{tang_unbiased_2020}} & VG150 & 18.4/25.4/28.7 & 8.9/12.2/14.0 & 6.9/9.3/11.1 & \bf - \\
         & VG150-con & 22.5/31.22/37.02 & 9.38/13.32/15.29 & 8.56/10.84/13.09 & \bf $\uparrow$ 19.5 \%\\
         & VG150-cur & 22.03/32.25/38.24 & 13.73/18.14/20.70 & 10.89/14.52/17.09 & \bf $\uparrow$ 39\% \\
         \hline
    \end{tabular}
    \caption{Reported performance of baseline models on different datasets, * denotes results reproduced using code by the authors. Improvements are the relative average against the baseline VG150.}
    \label{tab:results}
\end{table*}

\section{Experimental Setup and Results}\label{expe}

To demonstrate that the VG150 benchmark is biased and does not correctly evaluate the performance of approaches in the task, we conducted experiments with baseline SGG models. We follow previous work in the area \cite{xu2017scene, zellers2018neural, tang2019learning, tang_unbiased_2020} by evaluating our approach on three distinct (but related) tasks, namely Predicate Classification \textit{PredCls}, Scene Graph Classification \textit{SGCls}, and Scene Graph Generation \textit{SGGen}. \textit{PredCls} concentrates on predicting a relation, given the bounding boxes and \(<subject, object>\) pairs. \textit{SGCls} is analogous to \textit{PredCls}, except that \(<subject, object>\) pairs are not known {\em a priori} and they need to be inferred by the model. Finally, \textit{SGGen} assumes no prior knowledge; thus, the task included the prediction of object regions, pairs, and relations. To be consistent with other related work, a selection of the most used baseline models were trained: IMP \cite{xu2017scene}, Motifs \cite{zellers2018neural}, and VCTree \cite{tang2019learning}. For Motifs and VCTree, we trained the TDE version introduced in \cite{tang_unbiased_2020}. 
As other metrics have proven to be ineffective to measure the performance for both the head and tail classes \cite{tang2019learning}, we used the meanRecall@K metric introduced in \cite{tang2019learning}. We trained the models on two distinct datasets: \textbf{(1)} a highly connected version of Visual Genome, VG150-con, where the goal was to evaluate the impact on the performance of the models with this highly-connected dataset; \textbf{(2)} the curated version of VG150 (proposed in this paper) where we removed all part-whole relations, as described in \Cref{filter_part_whole}, we call it VG150-cur. This last version represents a highly-connected data split with visually relevant annotations. 

We retrained every model using the code provided by the authors \cite{tang_unbiased_2020}\footnote{\scriptsize{\url{https://github.com/KaihuaTang/Scene-Graph-Benchmark.pytorch}}}, whereby the original parameters were maintained, except for the batch size and learning rate that were fit to our hardware requirements. The Faster-RCNN backbone was trained using the same configuration as \cite{tang_unbiased_2020}, and we ensured that the mAp values were similar to that reported in the original paper, in order to guarantee a fair comparison in the SGGen task (respectively 0.24 and 0.27 for VG150-con and VG150-cur, whereas the original Faster-RCNN trained on VG150 has a mAp of 0.28). The training was conducted with a batch size of 32 and a base learning rate of 0.02 on one Nvidia RTX3090 within 20000 iterations (approximately 10 epochs) or 30000 iterations for SGGen. IMP was retrained on the baseline split (VG150) with the above settings, this is why the reported results in \Cref{tab:results} are slightly different from those reported in the original paper \cite{tang_unbiased_2020}. For comparison, the same training/validation/test split of the original VG150 was used for all datasets.

\subsection{Quantitative Results}
The results obtained with the experiments conducted in this work are listed in \Cref{tab:results}, where we can see that there was an improvement using VG150-con on the different baseline models (cf. the $6^{th}$ column of \Cref{tab:results}: {\em Improv.}). 
Neural Motifs and VCTree were the two models that benefited the most from the higher connectivity of the dataset for all the tasks. When we compared the statistics on the different data splits in \Cref{tab:statistics}, we noticed an improvement of 28.3\% in the number of the relation samples (train and test splits combined) and 9\% in the average vertex degree. This surely benefited the context learning of both Motifs and VCTree. The strong gap in performance between VG150 and VG150-con in PredCls also shows that context learning of baseline models is currently under-exploited.
Regarding the performance of the different models trained with VG150-curated, we observed a net improvement in the different tasks in contrast to the results obtained with VG150 and VG150-connected. In particular, there was an improvement of up to $39\%$ for VCTree-TDE model. We believe that the Total Direct Effect (TDE) strategy \cite{tang_unbiased_2020} highly benefited from the removal of irrelevant relationships as these are \textit{invariant} and, thus, biasing the reasoning process employed by the TDE. More concerning, we observe that Motifs-TDE performs best on average on the bias dataset VG150 but worse than VCTree-TDE on our non-bias version VG150-cur. This could impact the current ranking of SGG models on unbiased benchmarks such as VG150-cur.
Finally, we also noticed that VG150-cur is very similar in size to VG150 (see \Cref{tab:statistics}), however, VG150-cur possesses significantly more different triplets, and is thus more challenging to learn from. This result shows that, due to the removal of the invariant relations, the models were not biased into predicting invariant relations with over-confidence, improving by a high margin the meanRecall@K performance. 

\subsection{Qualitative Results}
\begin{figure*}[http]
     \centering
        \subfigure[]{
            \includegraphics[width=0.32\textwidth, height=12em]{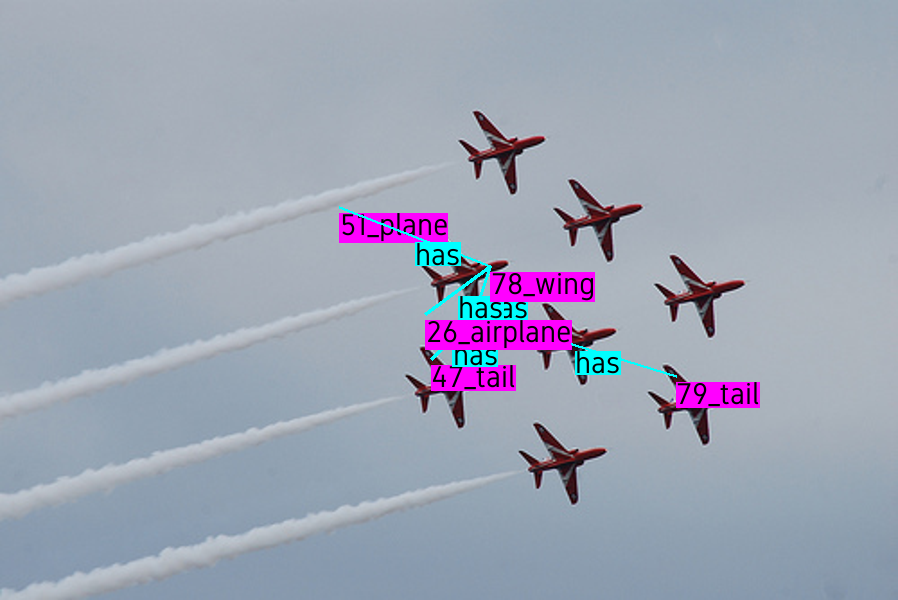}\label{fig:plane_baseline}
            } 
        \hfill
        \subfigure[]{
            \includegraphics[width=0.32\textwidth, height=12em]{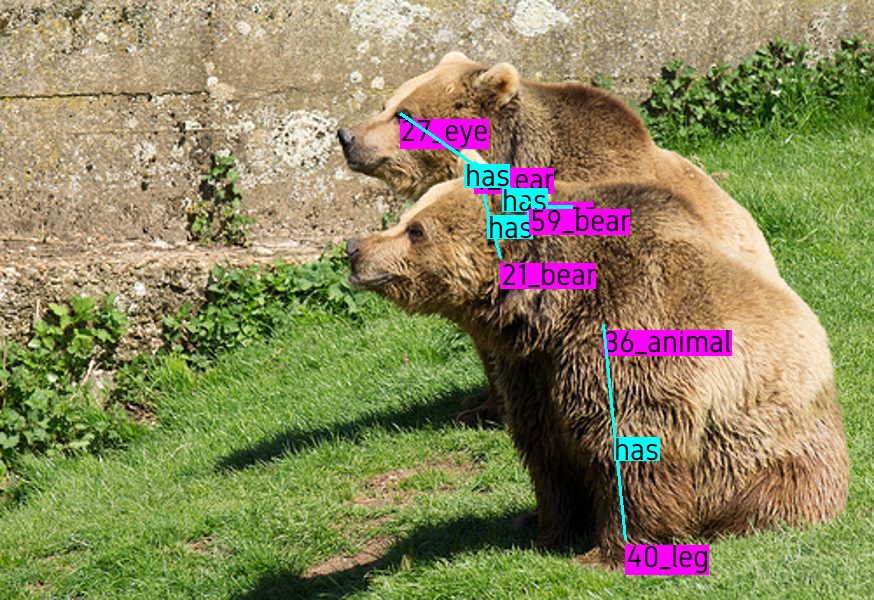}\label{fig:bear_baseline}
        } 
        \hfill
        \subfigure[]{
            \includegraphics[width=0.32\textwidth, height=12em]{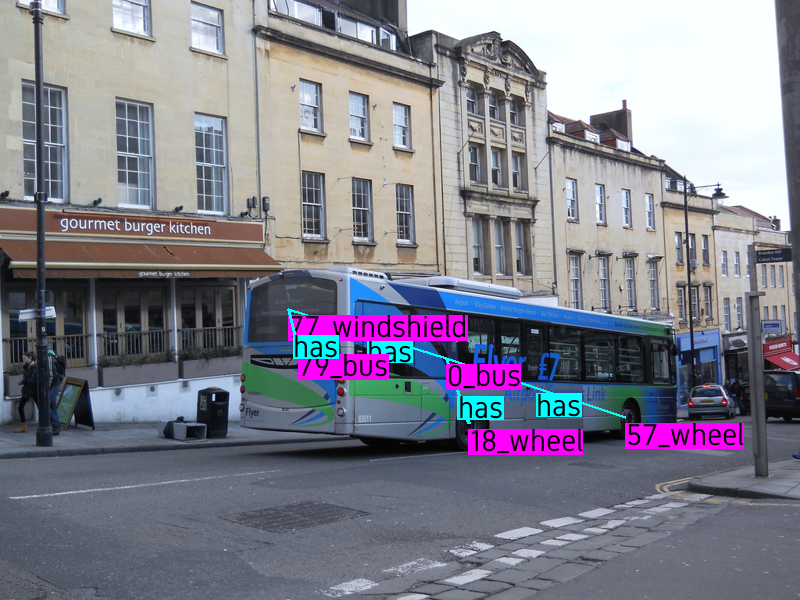}\label{fig:bus_baseline}
            } 
        \hfill
        \subfigure[]{
            \includegraphics[width=0.32\textwidth, height=12em]{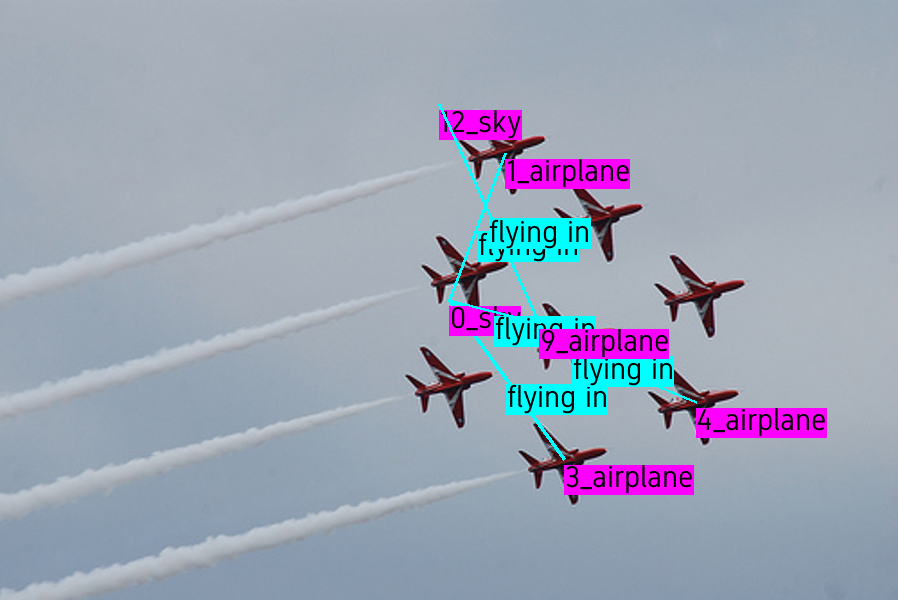}\label{fig:plane_curated}
        } 
        \hfill
        \subfigure[]{
            \includegraphics[width=0.32\textwidth, height=12em]{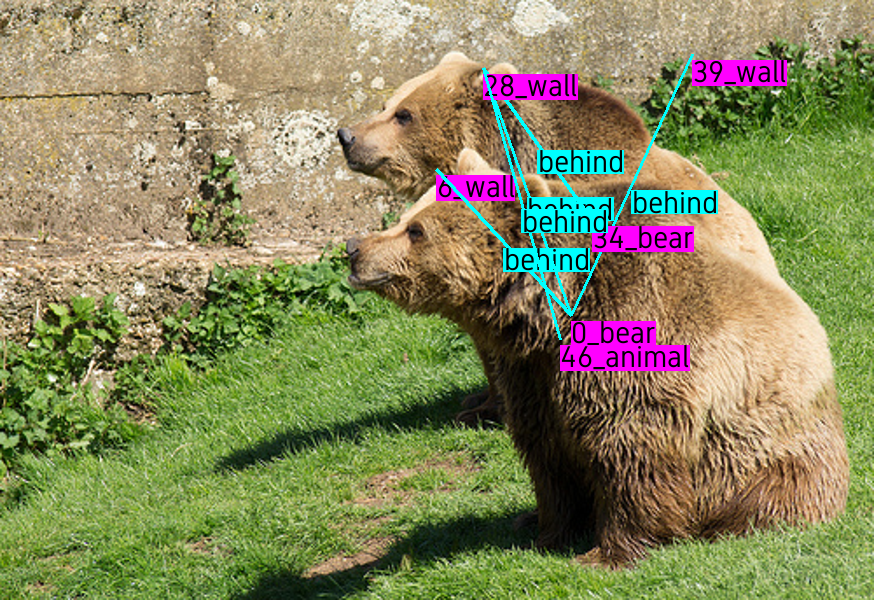}\label{fig:bear_curated}
            } 
        \hfill
        \subfigure[]{
            \includegraphics[width=0.32\textwidth, height=12em]{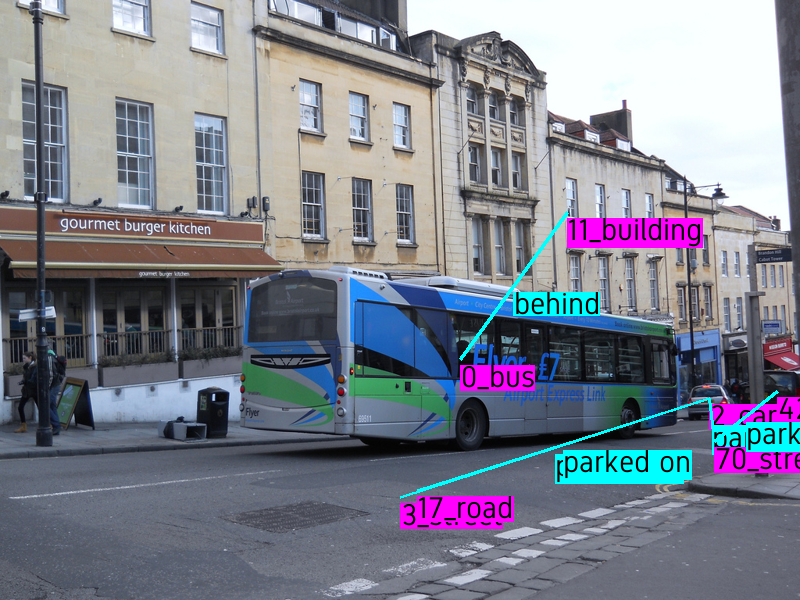}\label{fig:bus_curated}
        } 
    \caption{Top-5 predictions of Motifs-TDE \cite{tang_unbiased_2020} on the test set of the original VG150 dataset (top) and our curated version VG150-cur (bottom). Pink labels represent objects, and blue labels represent predicates. Best viewed in colour.}
    \label{fig:predictions}
\end{figure*}

In \Cref{fig:predictions} we compared predictions from the Motifs-TDE model \cite{tang_unbiased_2020} on the test set of the original VG150 and our curated version VG150-cur. On all images, we displayed the top 5 predicted relations. On \Cref{fig:plane_baseline,fig:bear_baseline,fig:bus_baseline} we can see that the main element of the image (planes, bears or a bus) are described through their internal components (\texttt{wing}, \texttt{tail} for the planes; \texttt{ear}, \texttt{eye} for the bears and \texttt{wheel}, \texttt{windshield} for the bus) whereas their interconnections with other elements from the image are missing. In the predictions made by training on VG150-curated, see \Cref{fig:plane_curated,fig:bear_curated,fig:bus_curated}, we can see that interactions with others elements (\texttt{sky}, \texttt{wall} and \texttt{road}, respectively) are present, giving more information about the scene. 
This example illustrates the problem with Visual Genome annotations and the bias in the learning process of SGG models: even if all predictions given by models trained on VG150 are correct, they are failing to provide useful information for the downstream tasks. The next section presents experiments on the task of Image Generation to illustrate this point.

\subsection{Evaluation}
\begin{figure}[]
     \centering
        \subfigure[]{
            \includegraphics[width=0.3\columnwidth]{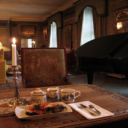}\label{fig:gan_gt}
            } 
        \hfill
        \subfigure[]{
            \includegraphics[width=0.3\columnwidth]{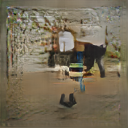}\label{fig:gan_baseline}
        } 
        \hfill
        \subfigure[]{
            \includegraphics[width=0.3\columnwidth]{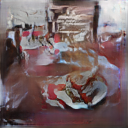}\label{fig:gan_curated}
            } 
        \hfill
    \caption{Images generated using sg2im \cite{johnson2018image}. Left: ground-truth image downsample in 128/128, middle: generation using graphs from VG150, right: generation with VG150-curated.}
    \label{fig:sg2im}
\end{figure}

As highlighted in \cite{tang_unbiased_2020}, the tasks of Visual Question Answering and Image Captioning rely on particular settings and external datasets with their own acknowledged biases. To remove those biases and outline the full potential of our approach, we chose to evaluate the quality of VG150-curated on the task of Image Generation from Scene Graphs \cite{johnson2018image}. In contrast to Image Captioning or VQA, Image Generation models can be trained directly from the raw dataset, without inputs of captions or question-answer pairs that could bias the evaluation. Thus, we used a straight-forward approach by retraining the popular Image Generation benchmark sg2im\footnote{\scriptsize{\url{https://github.com/google/sg2im}}} \cite{johnson2018image} with VG150-cur and VG150. We also compared it to the version of Visual Genome used by the original authors that possesses 178 object and 45 predicate classes \cite{johnson2018image}. We trained the model for 300,000 iterations with a batch size of 64 on one Nvidia RTX3090 GPU with a target image size of 128/128 pixels. 
To evaluate images generated with the different datasets, we use the Fréchet Inception Distance (FID) \cite{heusel2017gans}. In our case, this metric is evaluating the distance between the distribution of the ground truth images from the VG dataset and the one generated using the different graph annotations from VG150 and VG150-cur. We found out that this is the best metric to evaluate the quality of annotations because the more informative elements there are on the input graphs, the closest to the original image the generation should be. \Cref{tab:results_sg2im} shows our results obtained by retraining the model on the different datasets. We first observed that VG150 outperformed the Visual Genome split employed by the original authors by a strong margin. This is mainly due to the cleaning process of VG150 which is more elaborate than VG \cite{johnson2018image} (as described in \Cref{curation}). Then, we also noticed a strong improvement by using VG150-curated, this shows the clear benefit of our curation method for downstream tasks. In \Cref{fig:sg2im} we display a few generated samples, from where it is worth noting that the image generated using VG150 annotations (\ref{fig:gan_baseline}) is far from the target (\ref{fig:gan_gt}), whereas the version generated with the curated dataset proposed in this work (\ref{fig:gan_curated}) has more similar patterns with respect to \ref{fig:gan_gt}.

\begin{table}[]
    \centering
    \begin{tabular}{c|c}
        \hline
         \textbf{Dataset} & \textbf{FID Score $\downarrow$} \\
         \hline
         VG \cite{johnson2018image} & 143.2 \\ 
         VG150 & 115.2 \\
         VG150-curated & 96.8 \\
         \hline
    \end{tabular}
    \caption{Results on the Image Generation task using the Fréchet Inception Distance (FID), lower is better.}
    \label{tab:results_sg2im}
\end{table}

\section{Discussion}

The first work that used the Visual Genome dataset proposed a simple curation algorithm that selects relations only based on their overall frequency in the dataset \cite{xu2017scene}. 
Since then, this split has been used in the literature with no consideration of the type and interest of the annotation samples it contains. In this work, we showed that this approach was misleading and that Visual Genome contains more biases than the long-tail distribution of predicates \cite{tang_unbiased_2020}. 
We demonstrated that the over-representation of irrelevant relations in the training data leads the baseline models to predict useless relations with high confidence. By removing these relations, by the proposed new curation process, we observed an improvement of up to 39\% meanRecall@K on SGG baseline models, even if our new dataset is more diverse than VG150 (see \Cref{tab:statistics}). This work also showed the limit of the evaluation metric commonly used in the literature: as meanRecall@K is a ranking metric, its performance results can be easily biased by the prediction of invariant relations. These relations are also irrelevant for downstream tasks. The method proposed in this work was evaluated in the task of Image Generation where, with the use of our curated dataset, a better performance was achieved when compared to training the models on VG150. Finally, comparing SGG models on Image Generation has shown to be a more reliable way of comparing results, than the usual Recall or meanRecall metrics that are highly biased by the nature of the ground-truth annotations. 

\section{Conclusion}

In this work, we analysed two biases in the data distribution and annotations of the Visual Genome dataset. We then proposed two novel techniques to alleviate those biases, resulting in two new splits of the data, called VG150-connected and VG150-curated. These splits, in particular VG150-curated, facilitated an improvement in the SGG task using traditional evaluation metrics; thus, providing a fair comparison with respect to the most used dataset in the literature, VG150. We then analysed the obtained results qualitatively and quantitatively and demonstrated the correlation between higher-quality annotations and better representation learning. 
We hope this last point will help future approaches to obtain higher-quality datasets for the task. Future work will consider leveraging the principles of Efficient SGG for commonsense reasoning in embodied (robotics) agents. This shall include the use of spatial, semantic, and possessive relations for Visual Understanding in Human-Agent Open-Ended Interaction. 

{\small
\bibliographystyle{ieee_fullname}
\bibliography{SGG_biblio}

\begin{thebibliography}{10}\itemsep=-1pt

\bibitem{abdelkarim2021exploring}
Sherif Abdelkarim et~al.
\newblock Exploring long tail visual relationship recognition with large
  vocabulary.
\newblock In {\em Proceedings of the IEEE/CVF International Conference on
  Computer Vision}, pages 15921--15930, 2021.

\bibitem{abou2022topology}
David Abou~Chacra and John Zelek.
\newblock The topology and language of relationships in the visual genome
  dataset.
\newblock In {\em 2022 IEEE/CVF Conference on Computer Vision and Pattern
  Recognition Workshops (CVPRW)}, pages 4859--4867. IEEE, 2022.

\bibitem{amiri2022reasoning}
Saeid Amiri, Kishan Chandan, and Shiqi Zhang.
\newblock Reasoning with scene graphs for robot planning under partial
  observability.
\newblock {\em IEEE Robotics and Automation Letters}, 7(2):5560--5567, 2022.

\bibitem{amodeo2022og}
Fernando Amodeo et~al.
\newblock {OG-SGG}: Ontology-guided scene graph generation—a case study in
  transfer learning for telepresence robotics.
\newblock {\em IEEE Access}, 10:132564--132583, 2022.

\bibitem{cong_reltr_2022}
Yuren Cong, Michael~Ying Yang, and Bodo Rosenhahn.
\newblock {RelTR}: {Relation} {Transformer} for {Scene} {Graph} {Generation},
  Aug. 2022.
\newblock arXiv:2201.11460 [cs] version: 2.

\bibitem{dai_detecting_2017}
Bo Dai, Yuqi Zhang, and Dahua Lin.
\newblock Detecting {Visual} {Relationships} with {Deep} {Relational}
  {Networks}.
\newblock In {\em 2017 {IEEE} {Conference} on {Computer} {Vision} and {Pattern}
  {Recognition} ({CVPR})}, pages 3298--3308, Honolulu, HI, July 2017. IEEE.

\bibitem{damodaran2021understanding}
Vinay Damodaran et~al.
\newblock Understanding the role of scene graphs in visual question answering.
\newblock {\em arXiv preprint arXiv:2101.05479}, 2021.

\bibitem{desai2021learning}
Alakh Desai, Tz-Ying Wu, Subarna Tripathi, and Nuno Vasconcelos.
\newblock Learning of visual relations: The devil is in the tails.
\newblock In {\em Proceedings of the IEEE/CVF International Conference on
  Computer Vision}, pages 15404--15413, 2021.

\bibitem{dong_stacked_2022}
Xingning Dong et~al.
\newblock Stacked {Hybrid}-{Attention} and {Group} {Collaborative} {Learning}
  for {Unbiased} {Scene} {Graph} {Generation}.
\newblock In {\em 2022 {IEEE}/{CVF} {Conference} on {Computer} {Vision} and
  {Pattern} {Recognition} ({CVPR})}, pages 19405--19414, New Orleans, LA, USA,
  June 2022. IEEE.

\bibitem{gu2019scene}
Jiuxiang Gu, Handong Zhao, Zhe Lin, Sheng Li, Jianfei Cai, and Mingyang Ling.
\newblock Scene graph generation with external knowledge and image
  reconstruction.
\newblock In {\em Proceedings of the IEEE/CVF conference on computer vision and
  pattern recognition}, pages 1969--1978, 2019.

\bibitem{heusel2017gans}
Martin Heusel et~al.
\newblock Gans trained by a two time-scale update rule converge to a local nash
  equilibrium.
\newblock {\em Advances in neural information processing systems}, 30, 2017.

\bibitem{ilievski2021dimensions}
Filip Ilievski et~al.
\newblock Dimensions of commonsense knowledge.
\newblock {\em Knowledge-Based Systems}, 229:107347, 2021.

\bibitem{johnson2018image}
Justin Johnson, Agrim Gupta, and Li Fei-Fei.
\newblock Image generation from scene graphs.
\newblock In {\em Proceedings of the IEEE conference on computer vision and
  pattern recognition}, pages 1219--1228, 2018.

\bibitem{kan2021zero}
Xuan Kan, Hejie Cui, and Carl Yang.
\newblock Zero-shot scene graph relation prediction through commonsense
  knowledge integration.
\newblock In {\em Joint European Conference on Machine Learning and Knowledge
  Discovery in Databases}, pages 466--482. Springer, 2021.

\bibitem{krishna2017visual}
Ranjay Krishna et~al.
\newblock Visual genome: Connecting language and vision using crowdsourced
  dense image annotations.
\newblock {\em International journal of computer vision}, 123(1):32--73, 2017.

\bibitem{lee2019visual}
Soohyeong Lee, Ju-Whan Kim, Youngmin Oh, and Joo~Hyuk Jeon.
\newblock Visual question answering over scene graph.
\newblock In {\em 2019 First International Conference on Graph Computing (GC)},
  pages 45--50. IEEE, 2019.

\bibitem{li2022devil}
Lin Li, Long Chen, Yifeng Huang, Zhimeng Zhang, Songyang Zhang, and Jun Xiao.
\newblock The devil is in the labels: Noisy label correction for robust scene
  graph generation.
\newblock In {\em Proceedings of the IEEE/CVF Conference on Computer Vision and
  Pattern Recognition}, pages 18869--18878, 2022.

\bibitem{li2022sgtr}
Rongjie Li, Songyang Zhang, and Xuming He.
\newblock Sgtr: End-to-end scene graph generation with transformer.
\newblock In {\em Proceedings of the IEEE/CVF Conference on Computer Vision and
  Pattern Recognition}, pages 19486--19496, 2022.

\bibitem{li_bipartite_2021}
Rongjie Li, Songyang Zhang, Bo Wan, and Xuming He.
\newblock Bipartite graph network with adaptive message passing for unbiased
  scene graph generation.
\newblock In {\em Proceedings of the IEEE/CVF Conference on Computer Vision and
  Pattern Recognition}, pages 11109--11119, 2021.

\bibitem{li2022ppdl}
Wei Li et~al.
\newblock {PPDL}: Predicate probability distribution based loss for unbiased
  scene graph generation.
\newblock In {\em Proceedings of the IEEE/CVF Conference on Computer Vision and
  Pattern Recognition}, pages 19447--19456, 2022.

\bibitem{li2022embodied}
Xinghang Li, Di Guo, Huaping Liu, and Fuchun Sun.
\newblock Embodied semantic scene graph generation.
\newblock In {\em Conference on Robot Learning}, pages 1585--1594. PMLR, 2022.

\bibitem{liang2019vrr}
Yuanzhi Liang, Yalong Bai, Wei Zhang, Xueming Qian, Li Zhu, and Tao Mei.
\newblock Vrr-vg: Refocusing visually-relevant relationships.
\newblock In {\em Proceedings of the IEEE/CVF International Conference on
  Computer Vision}, pages 10403--10412, 2019.

\bibitem{lin_gps-net_2020}
Xin Lin, Changxing Ding, Jinquan Zeng, and Dacheng Tao.
\newblock {GPS}-{Net}: {Graph} {Property} {Sensing} {Network} for {Scene}
  {Graph} {Generation}.
\newblock In {\em 2020 {IEEE}/{CVF} {Conference} on {Computer} {Vision} and
  {Pattern} {Recognition} ({CVPR})}, pages 3743--3752, Seattle, WA, USA, June
  2020. IEEE.

\bibitem{liu2021fully}
Hengyue Liu, Ning Yan, Masood Mortazavi, and Bir Bhanu.
\newblock Fully convolutional scene graph generation.
\newblock In {\em Proceedings of the IEEE/CVF Conference on Computer Vision and
  Pattern Recognition}, pages 11546--11556, 2021.

\bibitem{lu2021context}
Yichao Lu et~al.
\newblock Context-aware scene graph generation with seq2seq transformers.
\newblock In {\em Proceedings of the IEEE/CVF international conference on
  computer vision}, pages 15931--15941, 2021.

\bibitem{plesse2020focusing}
Fran{\c{c}}ois Plesse, Alexandru Ginsca, Bertrand Delezoide, and Francoise
  Preteux.
\newblock Focusing visual relation detection on relevant relations with prior
  potentials.
\newblock In {\em Proceedings of the IEEE/CVF Winter Conference on Applications
  of Computer Vision}, pages 2980--2989, 2020.

\bibitem{reimers-2019-sentence-bert}
Nils Reimers and Iryna Gurevych.
\newblock Sentence-bert: Sentence embeddings using siamese bert-networks.
\newblock In {\em Proceedings of the 2019 Conference on Empirical Methods in
  Natural Language Processing}. Association for Computational Linguistics, 11
  2019.

\bibitem{ren2015faster}
Shaoqing Ren, Kaiming He, Ross Girshick, and Jian Sun.
\newblock Faster r-cnn: Towards real-time object detection with region proposal
  networks.
\newblock {\em Advances in neural information processing systems}, 28, 2015.

\bibitem{speer2017conceptnet}
Robyn Speer, Joshua Chin, and Catherine Havasi.
\newblock Conceptnet 5.5: An open multilingual graph of general knowledge.
\newblock In {\em Thirty-first AAAI conference on artificial intelligence},
  2017.

\bibitem{suhail_energy-based_2021}
Mohammed Suhail, Abhay Mittal, Behjat Siddiquie, Chris Broaddus, Jayan Eledath,
  Gerard Medioni, and Leonid Sigal.
\newblock Energy-{Based} {Learning} for {Scene} {Graph} {Generation}.
\newblock In {\em 2021 {IEEE}/{CVF} {Conference} on {Computer} {Vision} and
  {Pattern} {Recognition} ({CVPR})}, pages 13931--13940, Nashville, TN, USA,
  June 2021. IEEE.

\bibitem{tang_unbiased_2020}
Kaihua Tang, Yulei Niu, Jianqiang Huang, Jiaxin Shi, and Hanwang Zhang.
\newblock Unbiased {Scene} {Graph} {Generation} {From} {Biased} {Training}.
\newblock In {\em 2020 {IEEE}/{CVF} {Conference} on {Computer} {Vision} and
  {Pattern} {Recognition} ({CVPR})}, pages 3713--3722, Seattle, WA, USA, June
  2020. IEEE.

\bibitem{tang2019learning}
Kaihua Tang, Hanwang Zhang, Baoyuan Wu, Wenhan Luo, and Wei Liu.
\newblock Learning to compose dynamic tree structures for visual contexts.
\newblock In {\em Proceedings of the IEEE/CVF conference on computer vision and
  pattern recognition}, pages 6619--6628, 2019.

\bibitem{wang2019role}
Dalin Wang, Daniel Beck, and Trevor Cohn.
\newblock On the role of scene graphs in image captioning.
\newblock In {\em Proceedings of the Beyond Vision and LANguage: inTEgrating
  Real-world kNowledge (LANTERN)}, pages 29--34, 2019.

\bibitem{wang_exploring_2019}
Wenbin Wang, Ruiping Wang, Shiguang Shan, and Xilin Chen.
\newblock Exploring {Context} and {Visual} {Pattern} of {Relationship} for
  {Scene} {Graph} {Generation}.
\newblock In {\em 2019 {IEEE}/{CVF} {Conference} on {Computer} {Vision} and
  {Pattern} {Recognition} ({CVPR})}, pages 8180--8189, Long Beach, CA, USA,
  June 2019. IEEE.

\bibitem{wang2020sketching}
Wenbin Wang, Ruiping Wang, Shiguang Shan, and Xilin Chen.
\newblock Sketching image gist: Human-mimetic hierarchical scene graph
  generation.
\newblock In {\em European Conference on Computer Vision}, pages 222--239.
  Springer, 2020.

\bibitem{wen_unbiased_2020}
Bin Wen, Jie Luo, Xianglong Liu, and Lei Huang.
\newblock Unbiased {Scene} {Graph} {Generation} via {Rich} and {Fair}
  {Semantic} {Extraction}, Feb. 2020.
\newblock arXiv:2002.00176 [cs].

\bibitem{xu2017scene}
Danfei Xu, Yuke Zhu, Christopher~B Choy, and Li Fei-Fei.
\newblock Scene graph generation by iterative message passing.
\newblock In {\em Proceedings of the IEEE conference on computer vision and
  pattern recognition}, pages 5410--5419, 2017.

\bibitem{xu2019scene}
Ning Xu, An-An Liu, Jing Liu, Weizhi Nie, and Yuting Su.
\newblock Scene graph captioner: Image captioning based on structural visual
  representation.
\newblock {\em Journal of Visual Communication and Image Representation},
  58:477--485, 2019.

\bibitem{yan2020pcpl}
Shaotian Yan, Chen Shen, Zhongming Jin, Jianqiang Huang, Rongxin Jiang, Yaowu
  Chen, and Xian-Sheng Hua.
\newblock Pcpl: Predicate-correlation perception learning for unbiased scene
  graph generation.
\newblock In {\em Proceedings of the 28th ACM International Conference on
  Multimedia}, pages 265--273, 2020.

\bibitem{yang2022reformer}
Xuewen Yang, Yingru Liu, and Xin Wang.
\newblock Reformer: The relational transformer for image captioning.
\newblock In {\em Proceedings of the 30th ACM International Conference on
  Multimedia}, pages 5398--5406, 2022.

\bibitem{ye_linguistic_2021}
Keren Ye and Adriana Kovashka.
\newblock Linguistic {Structures} as {Weak} {Supervision} for {Visual} {Scene}
  {Graph} {Generation}.
\newblock In {\em 2021 {IEEE}/{CVF} {Conference} on {Computer} {Vision} and
  {Pattern} {Recognition} ({CVPR})}, pages 8285--8295, Nashville, TN, USA, June
  2021. IEEE.

\bibitem{yoon_unbiased_2022}
Kanghoon Yoon, Kibum Kim, Jinyoung Moon, and Chanyoung Park.
\newblock Unbiased {Heterogeneous} {Scene} {Graph} {Generation} with
  {Relation}-aware {Message} {Passing} {Neural} {Network}, Dec. 2022.
\newblock arXiv:2212.00443 [cs] version: 1.

\bibitem{yu_cogtree_2021}
Jing Yu, Yuan Chai, Yujing Wang, Yue Hu, and Qi Wu.
\newblock {CogTree}: {Cognition} {Tree} {Loss} for {Unbiased} {Scene} {Graph}
  {Generation}.
\newblock In {\em Proceedings of the {Thirtieth} {International} {Joint}
  {Conference} on {Artificial} {Intelligence}}, pages 1274--1280. International
  Joint Conferences on Artificial Intelligence Organization, Aug. 2021.

\bibitem{yu2017visual}
Ruichi Yu, Ang Li, Vlad~I Morariu, and Larry~S Davis.
\newblock Visual relationship detection with internal and external linguistic
  knowledge distillation.
\newblock In {\em Proceedings of the IEEE international conference on computer
  vision}, pages 1974--1982, 2017.

\bibitem{zellers2018neural}
Rowan Zellers, Mark Yatskar, Sam Thomson, and Yejin Choi.
\newblock Neural motifs: Scene graph parsing with global context.
\newblock In {\em Proceedings of the IEEE conference on computer vision and
  pattern recognition}, pages 5831--5840, 2018.

\bibitem{zhang2022fine}
Ao Zhang, Yuan Yao, Qianyu Chen, Wei Ji, Zhiyuan Liu, Maosong Sun, and Tat-Seng
  Chua.
\newblock Fine-grained scene graph generation with data transfer.
\newblock In {\em Computer Vision--ECCV 2022: 17th European Conference, Tel
  Aviv, Israel, October 23--27, 2022, Proceedings, Part XXVII}, pages 409--424.
  Springer, 2022.

\bibitem{zhang2017visual}
Hanwang Zhang, Zawlin Kyaw, Shih-Fu Chang, and Tat-Seng Chua.
\newblock Visual translation embedding network for visual relation detection.
\newblock In {\em Proceedings of the IEEE conference on computer vision and
  pattern recognition}, pages 5532--5540, 2017.

\bibitem{zhang2019large}
Ji Zhang, Yannis Kalantidis, Marcus Rohrbach, Manohar Paluri, Ahmed Elgammal,
  and Mohamed Elhoseiny.
\newblock Large-scale visual relationship understanding.
\newblock In {\em Proceedings of the AAAI conference on artificial
  intelligence}, volume~33, pages 9185--9194, 2019.

\bibitem{zhong_learning_2021}
Yiwu Zhong, Jing Shi, Jianwei Yang, Chenliang Xu, and Yin Li.
\newblock Learning to {Generate} {Scene} {Graph} from {Natural} {Language}
  {Supervision}.
\newblock In {\em 2021 {IEEE}/{CVF} {International} {Conference} on {Computer}
  {Vision} ({ICCV})}, pages 1803--1814, Montreal, QC, Canada, Oct. 2021. IEEE.

\bibitem{zhu_scene_2022}
Guangming Zhu, Liang Zhang, Youliang Jiang, Yixuan Dang, Haoran Hou, Peiyi
  Shen, Mingtao Feng, Xia Zhao, Qiguang Miao, Syed Afaq~Ali Shah, and Mohammed
  Bennamoun.
\newblock Scene {Graph} {Generation}: {A} {Comprehensive} {Survey}, June 2022.

\end{thebibliography}
}


\section*{Supplementary material}

# Supplementary Materials for Fine-Grained is Too Coarse: A Novel Data-Centric Approach for Efficient Scene Graph Generation

Maëlic Neau[1,2]     Paulo E. Santos[1]     Anne-Gwenn Bosser[2]     Cédric Buche[2]

[1]College of Science and Engineering, Flinders University, Australia

[2]Ecole Nationale d'Ingénieurs de Brest, France

{neau, buche, bosser}@enib.fr, paulo.santos@flinders.edu.au

## A. Imbalance of Predicates

After removing invariant relations, we noticed a slightly better balance in the predicate distribution of VG150-cur. In Table 1, we compare the Imbalance Ratio (IR) [10] and Imbalance Degree (ID) [9] of the different splits of Visual Genome. IR compares the distribution between head and tail classes only while ID compares the normalized distance between the actual distribution and a perfect distribution across all classes. However, as highlighted in [12], this metric is highly dependent on the type of distance chosen as well as the number of minority classes. As neither of the Imbalance Ratio nor Imbalance Degree gives a full picture of the imbalance in multi-class distribution, we also measure the likelihood-ratio imbalance degree (LRID) [12] as follows:

$$LRID = -2 \sum_{c=1}^{C} n_c \log \frac{N}{C n_c} \quad (1)$$

Where $C$ is the number of unique classes, $n_c$ is the frequency of each class and $N$ is the perfect distribution. This metric tests the actual distribution against a complete balance distribution of the data, and is reported to be more accurate for multi-class problems [12]. From Table 1 we observe that the difference between head and tail classes is lower by a strong margin when looking at IR between VG150 and VG150-cur. However, when looking at ID and LRID, we see that the imbalance over the global distribution is slightly similar, with a small advantage for VG150-cur. As unbiased SGG models are very sensitive to the imbalance of the dataset, this small difference could be a factor of the global improvement observed by training baseline models on VG150-cur.

## B. Training Strategy

For training on the task of Scene Graph Generation, we kept the same ratio of Training and Test images as VG150 with 0.7 and 0.3, respectively. In contrast to VG150 which set a fixed amount of 5000 validation images, we decided

| Dataset | IR ↓ | ID ↓ | LRID ↓ |
|---|---|---|---|
| VG80K | 600,210 | 29,278 | 13.75 |
| VG150 | 6,549 | 40.7 | 2.99 |
| VrR-VG | 619 | 95.61 | 2.50 |
| VG150-con | 1,697 | 40.69 | 2.98 |
| VG150-cur | 1,319 | 39.68 | 2.79 |

Table 1: Imbalance distribution measurement for different splits of the Visual Genome dataset.

to split the Training set between Train and Validation sets such as the validation set will have a ratio of 0.05 of the total amount of annotated images. Furthermore, the training strategy of VG150 is to keep the same split for both the training of Faster-RCNN and the relations training. However, a consequent amount of images are skipped when training the relation head due to the fact that they only have bounding boxes annotations. To preserve the train/val/test ratio, we then created a second split with images that contains at least one relation per image. This split is used exclusively when training the relation head while the first split is used for the training of the object detection backbone, see Table 2.

| Dataset | Object Train/Val/Test | Relationship Train/Val/Test |
|---|---|---|
| VG150 | 68,538/5,000/31,876 | 68,538/5,000/31,876 |
| VG150-con | 68,375/5,386/32,033 | 61,982/5,021/28,398 |
| VG150-cur | 68,380/5,385/32,049 | 59,948/4,875/26,980 |

Table 2: Statistics of the different training splits.

## C. Advanced Statistics

### C.1. Class Similarity

To give a fair comparison between VG150 and our approach, we also computed the similarities between predicate and object class distribution. VG150-cur possesses 88%

1

of similar predicate classes to VG150 while VG150-con possesses 92% similarity. Regarding object classes, both VG150-cur and VG150-con possess a 77% similarity with VG150. These statistics show that a comparison between VG150 and our approach makes sense as the distribution of classes is very similar.

### C.2. Scene Graph Generation Datasets

Even if Visual Genome is the largest and most widely adopted dataset in SGG, other datasets have been used such as VRD [11], Open Image [6] or GQA [1]. Table 3 shows a comparison between these datasets and Visual Genome. The Scene Graph [3] and VRD datasets [8] are too small and do not contains enough samples for efficient learning. On the other hand, OpenImage is a large-scale dataset but with very sparse annotations (e.g. average graph size of 2.7558) and a small number of predicate classes. Finally, the GQA dataset [2] is a subset of Visual Genome with the addition of Question-Answer pairs. Regarding scene graph annotations, GQA sees the addition of *to the left* and *to the right* relations and some manual refinement compared to Visual Genome. In the end, GQA possesses only 112,148 relationships other than *to the left* and *to the right*, making it less diverse and more sparse than Visual Genome or VG150. The manual refinement also did not cover the removal of any irrelevant relation. Because GQA annotations are based on Visual Genome, we did not create a GQA-con and GQA-cur splits. However, our method could be easily applied to the GQA dataset to remove irrelevant relations.

| Dataset | Images | Obj. | Pred. | # Rel. | Irr. |
|---|---|---|---|---|---|
| Visual Genome [4] | 108,073 | 95,394 | 33,121 | 2,316,063 | Yes |
| Scene Graph [3] | 5,000 | 6,745 | 1,310 | 112,707 | No |
| VRD [8] | 5,000 | 100 | 70 | 37,993 | No |
| OpenImage [5] | 133,503 | 601 | 30 | 367,914 | No |
| GQA [2] | 85,638 | 1,703 | 310 | 471,614 | Yes |

Table 3: Comparison of different Scene Graph datasets, # Rel. displays the total number of relationships and Irr. outlines the presence of irrelevant relations.

### C.3. Connectivity

Apart from average node degree or average graph size, looking at the average density of graphs across the different datasets gives some insight about the connectivity and inter-dependencies of samples. In table 4 we compare the average graph density to the number of annotated images. We see that VG150-cur possesses a higher density than VG150 while having more images.

The average density is a good metric however it does not provide a clear overview of inter-dependencies. In fact, all the relations on an image are not always dependent on each other (for instance between foreground/background regions). Thus, the average node degree metric used in the paper is more reliable. Table 4 also highlights the strong sparsity of VrR-VG [7] discussed in the paper when looking at both the average graph size and average node degree. This makes the VrR-VG unsuitable to evaluate the performance of context-aware models.

### D. Annotations Comparison

Figure 1 gives an overview of the difference in annotations between the original dataset, VG150, VG150-connected, and VG150-curated.

## References

[1] Xingning Dong et al. Stacked Hybrid-Attention and Group Collaborative Learning for Unbiased Scene Graph Generation. In *2022 IEEE/CVF Conference on Computer Vision and Pattern Recognition (CVPR)*, pages 19405–19414, New Orleans, LA, USA, June 2022. IEEE. 2

[2] Drew A Hudson and Christopher D Manning. Gqa: A new dataset for real-world visual reasoning and compositional question answering. In *Proceedings of the IEEE/CVF conference on computer vision and pattern recognition*, pages 6700–6709, 2019. 2

[3] Justin Johnson, Ranjay Krishna, Michael Stark, Li-Jia Li, David A. Shamma, Michael S. Bernstein, and Li Fei-Fei. Image retrieval using scene graphs. In *2015 IEEE Conference on Computer Vision and Pattern Recognition (CVPR)*, pages 3668–3678, Boston, MA, USA, June 2015. IEEE. 2

[4] Ranjay Krishna et al. Visual genome: Connecting language and vision using crowdsourced dense image annotations. *International journal of computer vision*, 123(1):32–73, 2017. 2

[5] Alina Kuznetsova, Hassan Rom, Neil Alldrin, Jasper Uijlings, Ivan Krasin, Jordi Pont-Tuset, Shahab Kamali, Stefan Popov, Matteo Malloci, Alexander Kolesnikov, Tom Duerig, and Vittorio Ferrari. The open images dataset v4: Unified image classification, object detection, and visual relationship detection at scale. *IJCV*, 2020. 2

[6] Rongjie Li, Songyang Zhang, and Xuming He. Sgtr: End-to-end scene graph generation with transformer. In *Proceedings of the IEEE/CVF Conference on Computer Vision and Pattern Recognition*, pages 19486–19496, 2022. 2

[7] Yuanzhi Liang, Yalong Bai, Wei Zhang, Xueming Qian, Li Zhu, and Tao Mei. Vrr-vg: Refocusing visually-relevant relationships. In *Proceedings of the IEEE/CVF International Conference on Computer Vision*, pages 10403–10412, 2019. 2

[8] Cewu Lu, Ranjay Krishna, Michael Bernstein, and Li Fei-Fei. Visual relationship detection with language priors. In *European Conference on Computer Vision*, 2016. 2

[9] Jonathan Ortigosa-Hernández, Inaki Inza, and Jose A Lozano. Measuring the class-imbalance extent of multi-class problems. *Pattern Recognition Letters*, 98:32–38, 2017. 1

[10] Ronaldo C Prati, Gustavo EAPA Batista, and Diego F Silva. Class imbalance revisited: a new experimental setup to assess the performance of treatment methods. *Knowledge and Information Systems*, 45:247–270, 2015. 1

| Dataset | Graph Size | Node Degree | Subgraphs | Subgraph Size | Density | # Images |
|---|---|---|---|---|---|---|
| VG80K | 19.0267 | 2.3474 | 3.4352 | 6.0786 | 0.1348 | 104,832 |
| VG150 | 6.9834 | 2.0262 | 1.9719 | 3.8775 | 0.2965 | 89,168 |
| VrR-VG | 3.1372 | 1.5547 | 1.5368 | 2.6451 | 0.3873 | 56,254 |
| VG150-con | 8.3794 | 2.1909 | 2.1018 | 4.1092 | 0.2752 | 95,400 |
| VG150-cur | 7.1453 | 2.1248 | 1.9487 | 3.6624 | 0.3219 | 91,803 |

Table 4: Comparison of connectivity and number of images in the different datasets. # Images represent the number of annotated images with at least one relation, different than the global number of annotated images as some images have only bounding boxes annotations.

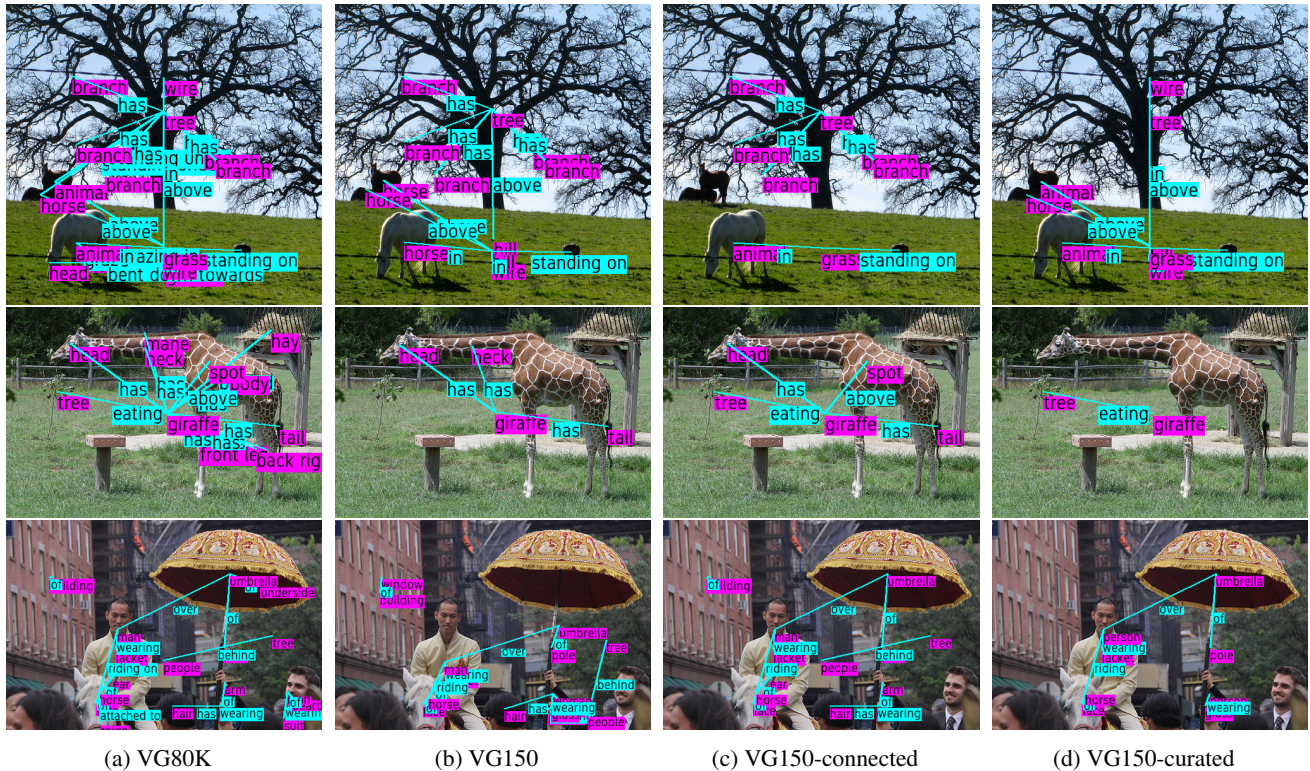

(a) VG80K     (b) VG150     (c) VG150-connected     (d) VG150-curated

Figure 1: A few examples of the difference between annotations in the original dataset VG80K, VG150, VG150-connected, and VG150-curated. We can easily see that annotation from VG150-curated (right) only detail informative relations, while irrelevant annotations are heavily present in the other data splits.

[11] Ruichi Yu, Ang Li, Vlad I Morariu, and Larry S Davis. Visual relationship detection with internal and external linguistic knowledge distillation. In *Proceedings of the IEEE international conference on computer vision*, pages 1974–1982, 2017. 2

[12] Rui Zhu, Ziyu Wang, Zhanyu Ma, Guijin Wang, and Jing-Hao Xue. Lrid: A new metric of multi-class imbalance degree based on likelihood-ratio test. *Pattern Recognition Letters*, 116:36–42, 2018. 1



\end{document}